# PRAGMATIC APPROACH TO STRUCTURED DATA QUERYING VIA NATURAL LANGUAGE INTERFACE

Aliaksei Vertsel, Mikhail Rumiantsau
FriendlyData, Inc.
San Francisco, CA
hello@friendlydata.io

## INTRODUCTION

As the use of technology increases and data analysis becomes integral in many businesses, the ability to quickly access and interpret data has become more important than ever. Information retrieval technologies are being utilized by organizations and companies to manage their information systems and processes. Despite information retrieval of a large amount of data being efficient organized in relational databases, a user still needs to master the DB language/schema to completely formulate the queries. This puts a burden on organizations and companies to hire employees that are proficient in DB languages/schemas to formulate queries. To reduce some of the burden on already overstretched data teams, many organizations are looking for tools that allow non-developers to query their databases.

Unfortunately, writing a valid SQL query that answers the question a user is trying to ask isn't always easy. Even seemingly simple questions, like "Which start-up companies received more than $200M in funding?" can actually be very hard to answer, let alone convert into a SQL query. How do you define start-up companies? By size, location, duration of time they have been incorporated?

This may be fine if a user is working with a database they're already familiar with, but what if users are not familiar with the database. What is needed is a centralized system that can effectively translate natural language queries into specific database queries for different customer database types. There is a number of factors that can dramatically affect the system architecture and the set of algorithms used to translate NL queries into a structured query representation.

## CLOUD BASED API VS. ON-PREMISE INSTALLED SYSTEMS

In a typical business case, internal data collected by corporations is presented in some structured form using standard DB storage platforms. In most cases data users already have a set of data access tools that they are accustomed to and NL driven system is viewed as an important convenience feature integrated a top of the existing data platform.

In this case, an API based cloud solution might seem the best choice as per time of integration and support. Yet a challenge arises: the external system needs to have access to internal corporate data which can compromise the data security.

The general rule applied in such business case is: the more access to data is granted to the cloud based system, the better quality of query translation can be provided. Basically we can identify 3 levels of data access:

1. Access to data schema (names/types of tables and columns and connections between them)
2. Access to dataset metadata (unique values of columns, e.g. names of organizations, people, products and categories)
3. Access to full DB records

When a business case requires both high quality query translation and high security standards of architecture, on-premise installed system may be the only possible solution.

## MACHINE LEARNING VS. RULE BASED QUERY ANALYSIS

Natural language queries tend to be a lot more diverse in their structure and wording than structured query languages. In real life use cases one can expect that users will enter grammatically incorrect questions, search engine like queries (string of search terms with no delimiters), use all kinds of synonyms and phrasing (usually industry slang). All that language needs to be parsed by the system and mapped to database entity names one way or another.

One approach to train the system is machine learning which requires a set of sample queries along with their translations into a structured form. This technique provides potentially great flexibility and can be easily extended by added new queries to the training set. There some serious drawbacks though: in real life situation, the users need the system to be fully functional at the first day of installation, long setup period associated with sample queries and corresponding translations collection doesn't seem to be an option.

The other approach is to use a grammar based on a set of manually constructed rules describing the basics of English sentence (grammatically correct language as well as partially correct and non-structured). Such rules can incorporate dictionaries of synonyms, metadata lists, mappings of custom phrases to structured language components.

In most cases, a hybrid approach combining the two worlds seems to be the optimal solution. The initial version of the system is based mainly on manually curated rules and dictionaries. As the user queries log starts to build up, some elements of machine learning may come into play improving the quality and giving more flexibility in phrasing.

### DATASET STRUCTURE AND OPERATIONS COMPLEXITY

Number of tables and connections in a dataset also affects the choice of the optimal query translation algorithm. Generally, machine learning approach is more applicable to simple (one table) structures with diverse query phrasing while rule based approach shows better results on multi-table datasets where not only mapping of natural language phrases to formal clauses is important but also hierarchy of nested structures.

A similar rule also works for simple vs. complex data operations: machine learning based algorithms show good results with basic sorting and aggregations while rule based or hybrid approach works more robustly with nested grouping, time series calculations, functions with multiple parameters etc.

### WIKISQL DATASET

When choosing the optimal solution for WikiSQL dataset question answering, we kept in mind the following factors:

- WikiSQL dataset contains thousands of unique schemas related to diverse domains which makes manual synonyms enrichment virtually impossible
- every test question always addresses one known table
- metadata is fully accessible
- metadata mentioned in test queries always keeps its original form (we can assume it was entered using an autosuggest widget)
- date formats in questions and table data might be different, yet limited to several distinct variants
- query object used as the output model allows the following operations: (1) filtering by multiple fields (equals, more, less) with "AND" operation only, (2) common aggregations (sum, count, min, max)

Though a training set is provided along with the WikiSQL data, we decided to emulate a more realistic business case where we have access to DB schema, metadata and no initial annotated query log.

QUESTION ANSWERING ALGORITHM DESCRIPTION

The basic hypothesis is that we can recognize and translate formal query structure by finding column and metadata in the input string and analyzing their sequence. Since the form of column names can be changed in the input, we apply partial column and metadata match applying the following rules:

**Use wildcards of 4 first letters instead of full names.**

**Is the word (first 4 letters) found in other columns?**
  Yes: Use this word as the column identifier
  No: Skip this word as it is ambiguous

**Is metadata (table cell value) found in this column only?**
  Yes: This cell value identifies both column and value
  No: This cell value needs to be match along with the column only

The schema and metadata are then incorporated into the generic grammar that defines the following logic:
- The query should contain the question word "what", "how many" or one of their synonyms
- Filtering conditions are expressed either by metadata values alone (for unique metadata values) or by metadata-column pairs
- Comparison operations are expressed by simple phrases following or preceding the columns (like "more than", "less than")
- Aggregation operations are expressed by words "minimal", "average", etc. and their synonyms
- Any number of other words is allowed between the words matched on the above stages

Let's consider the following schema:

| Year | Division | League | Regular Season | Playoffs | U.S. Open Cup |
|------|----------|--------|----------------|----------|---------------|
| real | text | text | text | text | text |
| 2002 | 3 | USISL | 4th, Atlantic Division | Divisional Semifinals | Did Not Qualify |
| 2003 | 3 | USISL | 9th, Atlantic Division | Did Not Qualify | Did Not Enter |

Figure 1. A Sample Table

The following constructs are identified in the input question:

{ how many } [ divisions ] { did not qualify } for [ u.s. open cup ] in { 2003 }

In this case "how many" is matched as the question word denoting the count aggregation, "divisions" is matched as the focus column. "Did not qualify" is a non-unique metadata item as it is present in two columns but we can identify the column it belongs to ("u.s. open cup") as it's explicitly present in the question. "2003" is matched as a unique metadata item, i.e. it implicitly and unambiguously identifies its column "year".

Based on the above matches and their sequence, we can form the following query object:
- aggregation: count
- selection column: division
- conditions: U.S. Open Cup = did not qualify AND Year = 2003

Below are a few more examples of questions that went through the same schema/metadata markup.

From what [ school ] was the { linebacker } that had a [ pick ] { less than } { 245 } and was drafted in [ round ] { 6 } ?

{ How many } [ artists ] were there for the [ show ] { thoroughly modern millie } ?

When { Naomi Owen } won the [ Women Singles ] and { Ricardo Walther } won the [ Men Singles ], { who } won the [ mixed veteran ] ?

EXAMPLE OF GRAMMAR IN BNF FORMAT

The matching algorithm is implemented using the proprietary FriendlyData Inc. parser driven by an FSM grammar declared in a classic BNF format with some extensions. Below we show a simplified fragment of such grammar and explain the its concepts as comments.

```
# definitions for columns "years in Toronto" and "school/club team"
# non-terminals are all caps, terminals are either all smalls for full words or /.../ for regex patterns

COLUMN_1@ ::= /year.*/ /in.*/ /toro.*/ | /year.*/ | /toro.*/
COLUMN_2@ ::= /scho.*/ /club.*/ /team.*/  |  /scho.*/  |  /club.*/  |  /team.*/
COLUMN ::= COLUMN_1 | COLUMN_2
# definitions of terminals can have a "@" extension denoting that a node with such name should be created once the pattern is matched
# unique metadata nodes don't need explicit presence of a column for disambiguation
UNIQ_METADATA_1@ ::= 2012 present | 1999 2000 | 1996 97 | 2002 03
UNIQ_METADATA_2@ ::= cincinnati | villanova | minnesota | detroit
UNIQ_METADATA ::= UNIQ_METADATA_1 | UNIQ_METADATA_2

# "& ~" construction means AND NOT and allows to add constraints and exceptions for patterns
UNKNOWN_WORD ::= /.+/ & ~( COLUMN | UNIQ_MD )

# "?" and "??" defines optional nodes. The former is greedy optionality (match if possible), the latter is non-greedy optionality (skip if possible)
```

```
UNKNOWN_WORD_CHAIN ::= UNKNOWN_WORD UNKNOWN_WORD ?? UNKNOWN_WORD
?? UNKNOWN_WORD ??

METADATA_1@ ::= 2012 present | 1999 2000 | 1996 97 | 2002 03 |
1996
METADATA_2@ ::= cincinnati | villanova | minnesota | detroit |
iowa | duke

# clause definition combines metadata with the corresponding
column allowing some number of unknown words between then, the
node order is arbitrary
COLUMN_METADATA_CLAUSE@ ::=
  COLUMN_1 UNKNOWN_WORD_CHAIN METADATA_1
| COLUMN_2 UNKNOWN_WORD_CHAIN METADATA_2
| METADATA_1 UNKNOWN_WORD_CHAIN COLUMN_1
| METADATA_1 UNKNOWN_WORD_CHAIN COLUMN_1

CLAUSE ::= ( COLUMN_METADATA_CLAUSE | UNIQ_METADATA | COLUMN )
UNKNOWN_WORD_CHAIN ??

WHAT@ ::= what | which | who
HOW_MANY@ ::= how many | how much | number | count | total

QUERY@ ::= UNKNOWN_WORD_CHAIN ?? (WHAT | HOW_MANY ) CLAUSE
CLAUSE ? CLAUSE ? CLAUSE ?
```

### PROCESSING QUALITY METRICS

This query translation method gives accuracy of **64.7%** for full query objects and **74.6%** for conditions only. Annotated queries in the training corpus can be used to identify synonyms that can't be matched by partial word comparison to further improve the query translation quality.